# USO DE GSO COOPERATIVOS COM DECAIMENTO DE PESOS PARA OTIMIZAÇÃO DE REDES NEURAIS

**Danielle N. G. Silva e Teresa B. Ludermir**

Centro de Informática – Universidade Federal de Pernambuco (UFPE)
Av. Jornalista Anibal Fernandes, s/n, 50.740-560 – Recife – PE – Brazil
{dngs, tbl}@cin.ufpe.br

**Abstract –** Training of Artificial Neural Networks is a complex task of great importance in supervised learning problems. Evolutionary Algorithms are widely used as global optimization techniques and these approaches have been used for Artificial Neural Networks to perform various tasks. An optimization algorithm, called Group Search Optimizer (GSO), was proposed and inspired by the search behaviour of animals. In this article we present two new hybrid approaches: CGSO-Hk-WD and CGSO-Sk-WD. Cooperative GSOs are based on the divide-and-conquer paradigm, employing cooperative behaviour between GSO groups to improve the performance of the standard GSO. We also apply the weight decay strategy (WD, acronym for Weight Decay) to increase the generalizability of the networks. The results show that cooperative GSOs are able to achieve better performance than traditional GSO for classification problems in benchmark datasets such as Cancer, Diabetes, Ecoli and Glass datasets.

**Resumo –** Treinamento de Redes Neurais Artificiais é uma tarefa complexa de grande importância em problemas de aprendizado supervisionado. Algoritmos Evolutivos são amplamente utilizados como técnicas de otimização global e estas abordagens têm sido usadas para Redes Neurais Artificiais executar várias tarefas. Um algoritmo de otimização, chamado de Otimizador por Grupo de Busca (GSO, acrônimo de Group Search Optimizer), foi proposto e inspirado pelo comportamento de busca dos animais. Neste artigo apresentamos duas novas abordagens híbridas: CGSO-$H_k$-WD e CGSO-$S_k$-WD. Os GSO cooperativos são baseados no paradigma de dividir e conquistar, empregando o comportamento cooperativo entre os grupos GSOs para melhorar o desempenho do GSO padrão. Também aplicamos a estratégia de decaimento de peso (WD, acrônimo de Weight Decay) para aumentar o poder de generalização das redes. Os resultados mostram que os GSO cooperativos são capazes de conseguir melhor desempenho do que o GSO tradicional para problemas de classificação de conjuntos de dados de Câncer, Diabetes, Ecoli e Vidro.

**Palavras Chave –** Redes Neurais Artificiais, Sistemas Híbridos, Otimizador por Grupo de Busca.

## 1 Introdução

Redes Neurais Artificiais (RNAs) são conhecidas como aproximadores universais e modelos computacionais com características particulares, como a adaptabilidade, a capacidade de aprender por exemplos e a capacidade de organizar ou generalizar os dados.

O processo de treinamento da Multilayer Perceptron (MLPs) para problemas de classificação de padrões é composto por duas tarefas difíceis, a primeira é a seleção de uma arquitetura apropriada para o problema, e a segunda é o ajuste dos pesos das conexões da rede. Algoritmos de aprendizagem tradicionais, tais como a técnica baseada no gradiente, chamado Backpropagation (BP) e seu variante Levenberg-Marquardt (LM) têm sido amplamente utilizados na formação das MLPs, mas estas abordagens podem ficar presas em mínimos locais. Atualmente, pode-se dizer que a área da Inteligência Artificial está em um estágio onde o objetivo final é a integração de diferentes técnicas para construir sistemas mais robustos, os chamados Sistemas Inteligentes Híbridos (SIHs). A motivação para esses sistemas serem desenvolvidos dá-se ao fato de que eles possibilitam a resolução de problemas complexos que só poderiam ser resolvidos quando tratados por cada uma das suas sub-tarefas, e unir as vantagens e superar as limitações individuais de diferentes técnicas. Técnicas de pesquisa global, tais como Algoritmos Evolutivos (AEs), como Algoritmos Genéticos (AG) [26], Busca Tabu (TS) [28], Otimização por Colônia de Formigas (ACO) [29] e Otimização por Enxame de Partículas (PSO, acrônimo de Particle Swarm Optimization) [11, 13], com a possibilidade de alargar o espaço de busca, na tentativa de evitar mínimos locais, tem sido amplamente combinados com Redes Neurais para realizar várias tarefas como: inicialização dos pesos das conexões, conexões dos pesos de treinamento, otimização da arquitetura, entre outros parâmetros da Rede Neural. Recentemente, uma nova técnica de otimização inspirada na natureza foi proposta: Otimização por Grupo de Pesquisa (GSO) [2, 3].

O algoritmo GSO é um novo algoritmo de Inteligência de Enxame (SI, acrônimo de Swarm Intelligence) para problemas de otimização. O algoritmo é baseado em um modelo de forrageiro social, Produtor-Dependente (PS, acrônimo de produtor-scrounger) modelo de [1], o que pressupõe que os membros do grupo de pesquisa ou "encontram" (producer) ou "juntam" (scrounging) oportunidades, diferentemente das metáforas usadas pelo ACO e PSO [7]. Algumas aplicações práticas têm sido feitas usando o algoritmo GSO. Em [4], a estratégia de otimização por grupo de pesquisa para treinamento de uma RNA foi utilizada para diagnóstico de câncer de mama. Em [5] foi apresentado um método de otimização multi-objetivo, onde o GSO com vários produtores (GSOMP) foi aplicado para o posicionamento ideal do sistema Flexible AC Transmission System (FACTS) para minimizar a perda real de energia e melhorar seu perfil de tensão. Em [6] foi aplicada uma RNA treinada com GSO para monitoramento da condição das máquinas de ultra-som. Todas essas aplicações usando o GSO têm consolidado com bons resultados a eficiência da sua abordagem.

Estratégias cooperativas que têm obtido bons resultados em problemas de otimização numérica são baseadas na divisão do espaço de busca. Assim, o espaço de busca original de dimensão n é dividido em 1 ≤ n ≤ K partições de dimensão d, com d = K × n [14, 16]. Em [23] foi introduzido esta idéia por meio de um algoritmo genético co-evolucionário (CCGA, acrônimo de coevolutionary genetic algorithm) para função de otimização, que consiste em uma decomposição de um problema n-dimensional (n variáveis), em problemas n 1-dimensional, cada um abordado por uma diferente sub-população (espécie). [24] apresentou um algoritmo para a integração de modelos substitutos da função de fitness com os procedimentos de busca coevolutivas. Em [14, 15], o modelo de Potter foi estendio, onde duas abordagens cooperativas foram apresentadas: CPSO-$S_K$ e CPSO-$H_K$. Neste artigo, apresentamos duas novas abordagens híbridas do GSO inspiradas nos estudos feitos dos artigos citados neste parágrafo, onde faz uso da cooperação entre os diferentes grupos do GSO. Em nosso trabalho adotamos 5 partições, combinadas com a heurística de decaimento de peso [17, 21] para melhorar o processo de otimização dos pesos das MLPs: A otimização por grupo de busca cooperativo (CGSO-$S_K$-WD, acrônimo de Cooperative Group Search Optimizer) e a otimização por grupo de busca híbrido cooperativo (CGSO-$H_K$-WD, acrônimo de Search Optimizer Hybrid Cooperative Group). O comportamento cooperativo é obtido pela estratégia de dividir e conquistar, onde cada grupo é responsável por um conjunto limitado de variáveis do problema, e a solução final é encontrada através da combinação de soluções encontradas por cada grupo do GSO. Estas técnicas são extensões dos trabalhos apresentados em [23] e [14, 15] para o contexto do algoritmo GSO. Os experimentos foram feitos com os algoritmos GSO e Levenberg-Marquardt tradicionais e nossas abordagens para problemas reais de classificação obtidos a partir do repositório UCI [32].

Este artigo está organizado em: Seção II apresenta o padrão de pesquisa do algoritmo GSO [2, 3], a heurística de decaimento de peso [17, 20] e as estratégias cooperativas. Em seguida, nossas abordagens de busca cooperativa usando o GSO são apresentadas na seção III e os resultados experimentais são mostrados na seção IV. As conclusões são dadas na Seção V.

## 2 Preliminares

### 2.1. Otimizador por Grupo de Busca (GSO)

O GSO [2, 3] é um algoritmo de enxame inteligente inspirado pelo comportamento social de pesquisa dos animais e na teoria de vida em grupo. O GSO emprega o modelo producer-scrounger (PS) [1]. Neste modelo, presume-se que há duas estratégias de forrageio em grupos: (1) produzindo, por exemplo, em busca de alimento, e (2) juntar (arrecadar), por exemplo, unindo recursos descobertos por outros. A população com tamanho N do GSO é chamada de *grupo* e cada indivíduo da população é chamado de *membro*. No espaço de busca n dimensional, o $i$th membro da $k$th iteração tem uma posição corrente $\mathbf{X}^k_i \in \Re^n$ e um ângulo de cabeça $\boldsymbol{\varphi}^k_i = (\varphi^k_{i\,1},..., \varphi^k_{i\,(n-1)}) \in \Re^{n-1}$. Cada direção de busca do $i$th membro, que é um vetor unitário $\mathbf{D}^k_i(\boldsymbol{\varphi}^k_i) = (d^k_{i1},..., d^k_{in}) \in \Re^n$ pode ser calculada a partir de $\boldsymbol{\varphi}^k_i$ como segue:

$$d^k_{i1} = \prod_{q=1}^{n-1} \cos(\varphi^k_{iq})$$
$$d^k_{ij} = \sin(\varphi^k_{i(j-1)}) \cdot \prod_{q=j}^{n-1} \cos(\varphi^k_{iq}) \quad (j=2,...,n-1) \quad (1)$$
$$d^k_{in} = \sin(\varphi^k_{i(n-1)}).$$

No GSO, um grupo é composto por três tipos de membros: os produtores, os dependentes e os membros dispersos [2, 3]. Existe apenas um produtor em cada busca do GSO [8] e os demais membros são dependentes ou membros dispersos. Todos os dependentes irão juntar o recurso encontrado pelo produtor. Durante cada iteração do GSO, um membro do grupo que encontrou o melhor valor de fitness é escolhido como produtor. Na iteração $k$th, o produtor $\mathbf{X}_p$ se comporta como se segue:

*Passo 1* - O produtor irá varrer a zero grau e lateralmente por três pontos no campo de digitalização da amostragem [7], um ponto a zero grau (2), um ponto no lado direito (3) e um ponto no lado esquerdo (4):

$$\mathbf{X}_z = \mathbf{X}^k_p + r_1 l_{max} \mathbf{D}^k_p(\boldsymbol{\varphi}^k) \quad (2)$$
$$\mathbf{X}_r = \mathbf{X}^k_p + r_1 l_{max} \mathbf{D}^k_p(\boldsymbol{\varphi}^k + \mathbf{r}_2 \theta_{max}/2) \quad (3)$$
$$\mathbf{X}_l = \mathbf{X}^k_p + r_1 l_{max} \mathbf{D}^k_p(\boldsymbol{\varphi}^k - \mathbf{r}_2 \theta_{max}/2) \quad (4)$$

onde $r_1 \in \Re^1$ é um número aleatório normalmente distribuído com média 0 e desvio padrão 1, $r_2 \in \Re^{n-1}$ é uma seqüência aleatória uniformemente distribuída no intervalo (0, 1), $\theta_{max} \in \Re^1$ é o ângulo de busca máxima, e $l_{max} \in \Re^1$ é a distância máxima de busca.

*Passo 2* - O produtor vai encontrar o melhor ponto (melhor valor de fitness) entre todos os pontos avaliados na etapa 1. Se o melhor ponto tem um valor melhor do que a sua posição atual, então ele vai voar para este ponto. Caso contrário ele vai ficar na sua posição atual e virar a cabeça para um novo ângulo gerado aleatoriamente:

$$\boldsymbol{\varphi}^{k+1}_i = \boldsymbol{\varphi}^k_i + \mathbf{r}_2 \alpha_{max} \quad (5)$$

onde $\alpha_{max} \in \Re^1$ é o ângulo máximo de viragem.

*Passo 3* - Se o produtor não pode achar uma melhor área após as iterações, ele vai virar a cabeça e voltar para zero grau:

$$\varphi^{k+a} = \varphi^k \quad (6)$$

onde $a \in \Re^1$ é uma constante dada por *round*($\sqrt{n+1}$) [3]. Em cada iteração, um número de membros do grupo é selecionado como dependentes, que continuam procurando oportunidades para juntar os recursos encontrados pelo produtor, adotando a estratégia de cópia de área [1]. Na iteração $k$th, o comportamento de copiar área do $i$th dependente, pode ser modelado como:

$$X_i^{k+1} = X_i^k + r_3 \circ (X_p^k - X_i^k) \quad (7)$$

onde $r_3 \in \Re^n$ é uma seqüência aleatória uniforme em (0, 1).

Os demais membros do grupo serão dispersos a partir de suas posições atuais. Caminhadas aleatórias são adotadas por esses membros dispersos como a estratégia adotada por [9]. Na iteração $k$th se o membro do grupo $i$th é selecionado como um membro disperso, ele gera um ângulo de cabeça aleatória $\varphi_i$ usando (5) e, em seguida, ele escolhe uma distância aleatória:

$$l_i = a \cdot r_1 l_{max} \quad (8)$$

e move para um novo ponto:

$$X_i^{k+1} = X_i^k + l_i D_i^k (\varphi^{k+1}) \quad (9)$$

No GSO, quando um membro está fora do espaço de busca, ele volta para sua posição anterior, dentro do espaço de busca, restringindo a pesquisa a um trecho rentável [10]. O algoritmo padrão do GSO está presente em [3].

## 2.2. Decaimento de Peso (WD)

Decaimento de peso foi inicialmente sugerido como uma heurística para melhorar o algoritmo de retropropagação das bias de uma robusta rede neural que é insensível ao ruído [17, 18, 21]. Esta aplicação usou um esquema adaptativo para a determinação do coeficiente de regularização ($\lambda(t)$), como descrito em [20]. Cada membro tem seu próprio coeficiente $\lambda_i(t)$ ajustado de forma adaptativa de acordo com seu erro médio. Após determinar a posição do indivíduo através da GSO padrão, o termo é aplicado à deterioração da posição atual do membro, de acordo com a equação 10.

$$X_i(t+1) = X_i(t+1) - \lambda_i(t+1) X_i(t+1) \quad (10)$$

A nova função de custo é dada por:

$$C = E_i(t) + \frac{\lambda_i(t)}{2} \|X_i\|^2 \quad (11)$$

Onde $E_i(t)$ denota o erro da rede neural no tempo $t$ relacionado ao indivíduo $i$ e $\|X_i\|^2$ é a norma ao quadrado de $X_i$. Em cada iteração, cada $\lambda_i(t)$ é atualizado da seguinte forma:

$$\lambda_i(t+1) = \begin{cases} \lambda_i(t) + INC, & \text{if } E_i(t) < \overline{E}_i(t) \\ \lambda_i(t) - INC, & \text{if } E_i(t) \geq \overline{E}_i(t) \\ \lambda_i(t), & \text{senão} \end{cases} \quad (12)$$

Onde *INC* é o valor do incremento do coeficiente de regularização e $\overline{E}_i(t)$ denota o erro médio obtido pelo indivíduo $i$th até um tempo $t$. O mecanismo de decaimento de peso atua em uma arquitetura de rede diferencialmente em direção a zero, reforçando grandes pesos e enfraquecendo conexões de pequenos pesos.

## 2.3. Abordagens Cooperativas

A cooperação envolve um conjunto de indivíduos que interagem através da comunicação de informação entre si, enquanto resolvem problemas [22]. Uma das primeiras formas de cooperação foi estudada com base na evolução das ilhas [25, 26], onde cada ilha corresponde a uma sub-população isolada, pesquisando em uma área separada do espaço de solução. Depois de $m$ iterações, as ilhas enviam e recebem $1 \leq z \leq 5$ indivíduos que promovem um intercâmbio de informações entre as ilhas. A forma de cooperação é baseada na abordagem celular [25], onde o conceito de vizinhança é usado.

Uma estratégia de cooperação que tem obtido melhores resultados em problemas de otimização numérica é a baseada na divisão do espaço de busca. Esta estratégia foi inicialmente introduzida por [23], por meio de um algoritmo genético co-evolucionário cooperativo (CCGA). A função de fitness para um membro da população é obtida pela combinação de suas variáveis com as melhores soluções encontradas até o momento pelas sub-populações restantes. No PSO esta abordagem foi introduzida pelo algoritmo CPSO-S$_K$ [14, 15].

## 3. Otimizador por Grupo de Busca Cooperativos

Esta seção apresenta as nossas duas novas abordagens híbridas do GSO baseadas na cooperação: CGSO-S$_K$-WD e CGSO-H$_K$-WD. Aplicamos também a heurística de decaimento de peso para aumentar o poder de generalização das nossas metodologias. Em nossa abordagem, adotamos o tratamento de absorção para os membros que escapam do espaço de busca [31]. Em CGSO-S$_K$-WD o espaço de dimensão $n$ da busca tem sido dividido em partições $K$ de dimensão $d$ em que uma sub-pesquisa é executada. Formalmente, a cooperação entre as sub-populações é feita através da concatenação dos sub-indivíduos atuais que desejam avaliar e os melhores sub-indivíduos até o momento, das demais partições, nas posições correspondentes. Esta composição é representada pelo contexto do vetor resultante da função $b(j, vec)$ expressa na equação (13), com $vec$ como o sub-individuo da sub-população $j$ que queremos avaliar e $G_i.X_p$ como o melhor sub-individuo da sub-população $i$.

$$b(j, vec) \equiv (G_1.X_p, \ldots, G_{j-1}.X_p, \ldots, G_{j+1}.X_p, \ldots, G_K.X_p) \quad (13)$$

Assim, podemos descrever o algoritmo CGSO-S$_K$ como sendo realizado pela cooperação de GSOs que otimiza cada uma das partições $K$ do espaço de busca. Nosso método CGSO-H$_K$ é baseado no algoritmo CPSO-H$_K$ apresentado em [14, 15]. A troca de informações é realizada de tal forma que alguns membros de uma metade da população do algoritmo são substituídos pela melhor solução descoberta até o momento pela outra metade do algoritmo. Depois de uma iteração da metade do algoritmo CGSO-S$_K$ (o grupo $G_j$), o vetor do contexto $b(1, G_1.X_p)$ é usado para substituir uma partícula escolhida aleatoriamente da metade do algoritmo GSO (o grupo $Q$). Isso é seguido por uma iteração do componente do grupo $Q$, que produz um novo global, $Q.X_p$. Esse vetor é então dividido em sub-vetores de dimensão apropriada e utilizado para substituir as posições das partículas escolhidas aleatoriamente nos grupos $G_j$. O algoritmo não substitui o $X_p$ de nenhum grupo [15]. Figura 1 e Figura 2 mostram os pseudocódigos do CGSO-S$_K$-WD e do CGSO-S$_K$-WD, respectivamente. Nessas abordagens, cada membro $i$ do grupo é composto por um conjunto de pesos entre a camada de entrada e nós ocultos ($W_1$), bias ocultas ($\Theta_1$), um conjunto de pesos entre a camada escondida e a camada de saída ($W_2$), e bias de saída ($\Theta_2$) [4]:

$$X_i = [W^i_1, \Theta^i_1, W^i_2, \Theta^i_2]$$

Para cada membro, todas as variáveis são inicializadas aleatoriamente dentro do intervalo [-1, 1]. A função de fitness adotada é o erro médio quadrático (MSE, acrônimo de mean squared error) no conjunto de validação:

$$E = \frac{1}{N} \sum_{n=1}^{N} \sum_{k=1}^{C} \left(t_k^n - o_k^n\right)^2 \quad (14)$$

Onde $t^n_k$ é a saída desejada relacionada ao indivíduo $i$th referente à classe $k$th, $o^n_k$ é a saída obtida pela RNA para o indivíduo $i$th referente à classe $k$th e $C$ é o número de nós de saída.

Sobre o algoritmo 1 e 2, podemos destacar que o valor de fitness é calculado através do erro da RNA, e o critério de parada adotado foi este erro chegar ao valor mínimo igual a 0, além do número máximo de iteração do GSO e da RNA.

## 4. Resultados e Discussões

Nesta seção, nós comparamos o tempo de treinamento e a accuracy de teste do GSO tradicional com o LM e nosso novo CGSO. Todos os programas são executados no MATLAB 6,0, e o algoritmo LM é fornecido pelo toolbox de redes neurais do MATLAB. Um conjunto de validação é usado em todas as metodologias avaliadas para evitar overfitting. Para avaliar todos estes algoritmos foram utilizados quatro conjuntos de dados de classificação: câncer, diabetes, vidro e ecoli, obtido a partir do repositório UCI [32]. Estes conjuntos de dados apresentam diferentes graus de dificuldades, números de classes diferentes e números de exemplos e características diferentes. Para estas abordagens, consideraram-se redes com apenas uma camada escondida, com seis neurônios nesta camada, realizando o ajuste dos pesos correspondentes.

As métricas de avaliação utilizadas são de uma análise empírica e análise de variância (ANOVA) que chama o valor de teste $F$ (teste $F$ formulado pelo BioEstat 5.0), complementada com o exame, a priori (Bonferroni), das diferenças entre as médias amostrais. Quando o valor de $F$ é significativo e a escolha do teste das diferenças entre as médias for de Bonferroni, deve ser escolhido previamente o nível alfa, ou seja, como o teste é bilateral temos ($\alpha/2$) e escolhemos alfa igual a 0.025.

Em nossos experimentos, alguns parâmetros foram estabelecidos para todos os conjuntos de dados (Tabela 1), de acordo com [2, 3, 15, 16]. A distância máxima busca $l_{max}$ foi dada como segue:

$$l_{max} = \|U - L\| = \sqrt{\sum_{i=1}^{n}(U_i - L_i)^2} \quad (15)$$

Onde Li e Ui são os limites inferiores e superiores para dimensão $i$th.

```
1.  k = 0;
2.  PARA (cada grupo j ∈ K)
3.      Inicialize aleatoriamente posições Gj.Xi ângulo de cabeça φi de todos os membros;
4.      Calcule o valor inicial da função de custo dos membros: f(Gj.Xi);
5.  FIM PARA
6.  ENQUANTO (Se as condições finais não forem atendidas)
7.      PARA (cada grupo j ∈ K)
8.          PARA (cada membro i do grupo)
9.              Escolher o produtor: encontre o produtor Xp do grupo;
10.             Performance do produtor: Comportamento do produtor como nos passos 1, 2 e 3;
11.             Performance do dependente: Selecione uma porcentagem dos restantes dos membros para ser
                dependentes;
12.             Performance dos membros dispersos: Os restantes dos membros irão executar a performance dos
                membros dispersos:
                    a.  Gerar o ângulo de cabeça aleatório usando (5);
                    b.  Escolha a distância li usando (8) e mova para o novo ponto usando (9);
13.             Avaliação do Fitness: Calcule o valor do fitness do membro atual: f(Xi);
14.             Atualização do coeficiente de regularização: Atualizar todos os decaimentos de pesos usando (12);
15.         FIM PARA
16.     FIM PARA
17. k = k + 1;
18. FIM ENQUANTO
```

**Figura 1 – Pseudocódigo do algoritmo CGSO-S$_k$-WD**

Os teste foram realizado para cada conjunto diferente de dados, número máximo de iterações para o GSO igual a 50, usando o melhor valor global (tabela 1). Para LM, o número máximo de iterações foi definido igual a 200 e o tamanho do grupo foi definido igual a 50 [3].

Tabela 1 – Parâmetros fixos para todos os algoritmos

| Algoritmo | Parâmetro | Valor |
|---|---|---|
| GSO | Porcentagem dos dependentes | 80% |
| | $\theta_{max}$ | $\pi/a^2$ |
| | $\alpha max$ | $\theta_{max}/2$ |
| | Número máximo de iterações | 50 |
| LM | Número de nós escondidos | 6 |
| | Número máximo de iterações | 200 |
| Decaimento de Pesos | $\Lambda$ | 5 X 10-6 |
| | INC | 1 X 10-3 |

Cada conjunto de dados foi dividido em conjuntos de validação, treinamento e testes, conforme especificado na Tabela 2. Para todos os algoritmos, 50 execuções independentes foram feitas com cada conjunto de dados. Os conjuntos de validação, treinamento e testes foram gerados aleatoriamente a cada tentativa de simulações. Os resultados obtidos pela LM, GSO e a abordagem GSO cooperativa que alcançaram os melhores resultados estão em negrito.

```
1.   k = 0;
2.   PARA (cada grupo j ∈ K)
3.       Inicialize aleatoriamente posições $G_j.\mathbf{X}_i$ e ângulos de cabeça $\varphi_i$ de todos os membros;
4.       Calcule o valor da função de custo iniciais dos membros: $f(G_j.\mathbf{X}_i)$;
5.   FIM PARA
6.   Inicialize o n-dimensional GSO: Q
7.   ENQUANTO (as condições finais não forem atendidas)
8.       PARA (cada grupo j ∈ K)
9.           PARA (cada membro i do grupo)
10.              Escolher produtor: Encontre o produtor $\mathbf{X}_p$ do grupo;
11.              Performance do produtor: Comportamento do produtor como nos passos 1, 2 e 3;
12.              Performance do dependente: Selecione uma porcentagem dos membros para executar a performance dos dependentes;
13.              Performance de dispersão: Os restantes dos membros irão dispersar:
                     a. Gerar um ângulo de cabeça aleatório usando (5);
                     b. Escolher a distância $l_i$ usando (8) e mova para um novo ponto usando (9);
14.              Evolução do Fitness: Calcule o calor da função de custo do membro atual: $f(\mathbf{X}_i)$;
15.              Atualizar coeficiente regularização: atualizar decaimentos de pesos usando (12);
16.          FIM PARA
17.      FIM PARA
18.      Selecione aleatoriamente $i \sim U(1, N/2 \mid Q.\mathbf{X}_i \neq Q.\mathbf{X}_p$
19.      $Q.\mathbf{X}_i = b(1, G_l.\mathbf{X}_p)$
20.      PARA (cada membro i do grupo)
21.          Escolher produtor: Encontre o produtor $\mathbf{X}_p$ do grupo;
22.          Performance do produtor: comportamento do produtor como nos passos 1, 2 e 3;
23.          Performance dos dependentes: Selecione uma porcentagem dos membros para executar a performance dos dependentes;
24.          Performance de dispersão: Os restantes dos membros irão dispersar:
                 b. Gerar um ângulo de cabeça aleatório usando (5);
                 c. Escolher a distância $l_i$ usando (8) e mova para um novo ponto usando (9);
25.          Evolução do Fitness: Calcule o valor da função de custo do membro atual: $f(\mathbf{X}_i)$;
26.          Atualizar coeficiente de regularização: Atualizar os decaimentos pesos usando (12);
27.      FIM PARA
28.      PARA cada grupo j ∈ [1…K]:
29.          Selecione aleatoriamente $i \sim U(1, s/2 \mid G_j.\mathbf{X}_i \neq G_j.\mathbf{X}_p$
30.          $G_j.\mathbf{X}_i = Q.\mathbf{X}_p$
31.      FIM PARA
32.      k = k + 1;
33.  FIM ENQUANTO
```

**Figura 2 – Pseudocódigo do algoritmo CGSO-$H_k$-WD**

Tabela 2 – Especificações das bases de dados

| Bases de Dados | Câncer | Diabetes | Ecoli | Vidros |
|---|---|---|---|---|
| Treinamento | 350 | 384 | 180 | 114 |
| Validação | 175 | 192 | 78 | 50 |
| Teste | 174 | 192 | 78 | 50 |

Na Tabela 3 a Tabela 6, os resultados obtidos para cada método são mostrados. A Tabela 3 mostra os resultados obtidos para a base câncer. O melhor resultado foi conseguido pelo CGSO-$H_k$-WD (96,40% de Accuracy de teste) em uma avaliação empírica, e todas as abordagens variantes do GSO superaram o GSO tradicional e o algoritmo LM. A análise de variância (95% intervalo de confiança) mostrou que CGSO-$H_k$-WD e CGSO-$S_k$-WD alcançaram melhores resultados do que o algoritmo GSO tradicional e LM. Apesar do tempo alcançado por eles ser um pouco maior do que tempo atingido pelas outras abordagens.

Tabela 3 – Resultados para Câncer

| Algoritmos | Tempo de Treinamento (s) | Accuracy de Teste (%) |
|---|---|---|
| LM | 0,28376 | 87,32 ± 15,22 |
| GSO | 1041,38 | 95,68 ± 1,44 |
| GSO-WD | 1043,84 | 96,08 ± 1,41 |
| GSO-$S_k$-WD | 1262,58 | 96,33 ± 1,67 |
| GSO-$H_k$-WD | 1249,74 | **96,40 ± 1,46** |

Para a base de diabetes (Tabela 4), em uma análise empírica, o CGSO-$S_k$-WD (76,42 de Accuracy de teste) superou todos os outros algoritmos, e todas as abordagens híbridas foram superiores ao GSO tradicional. O teste ANOVA mostrou que o algoritmo CGSO-$S_k$-WD foi melhor do que o GSO tradicional e o algoritmo LM.

Tabela 4 – Resultados para Diabetes

| Algoritmos | Tempo de Treinamento (s) | Accuracy de Teste (%) |
|---|---|---|
| LM | 0,31474 | 70,61 ± 7,99 |
| GSO | 1205,83 | 75,59 ± 2,39 |
| GSO-WD | 1246,65 | 75,61 ± 2,80 |
| GSO-$S_k$-WD | 1368,50 | **76,42 ± 2,38** |
| GSO-$H_k$-WD | 1585,08 | 76,23 ± 2,56 |

A tabela 5, para a classificação de tipos de vidros, o GSO-WD (82,74) obtive maior Accuracy de teste para uma análise empírica. Alguns dos algoritmos baseados no GSO superaram o GSO tradicional e LM empiricamente, mas de acordo com teste ANOVA seus resultados são semelhantes entre si. O GSO-$H_k$-WD apesar de apresentar o valor da accuracy de teste maior do que o GSO tradicional, não apresenta um tempo de treinamento melhor do que o GSO tradicional.

Tabela 5 – Resultados para Vidros

| Algoritmos | Tempo de Treinamento (s) | Accuracy de Teste (%) |
|---|---|---|
| LM | 1,1889 | 53,64 ± 23,50 |
| GSO | 2362,94 | 82,64 ± 3,52 |
| GSO-WD | 2841,36 | 81,82 ± 7,12 |
| GSO-$S_k$-WD | 3467,36 | 82,00 ± 5,77 |
| GSO-$H_k$-WD | 3604,1 | **82,74 ± 4,70** |

Em relação ao conjunto das bactérias Escherichia coli, conhecida como Ecoli, o melhor resultado empírico foi atingido por GSO-WD (64,24). Dois algoritmos baseado no GSO superaram o GSO tradicional e LM. O teste ANOVA mostrou que o GSO-WD foi melhor do que o GSO tradicional e o algoritmo LM.

Tabela 6 – Resultados para Ecoli

| Algoritmos | Tempo de Treinamento (s) | Accuracy de Teste (%) |
|---|---|---|
| LM | 0,50688 | 45,96 ± 16,80 |
| GSO | 1917,85 | 61,96 ± 6,17 |
| GSO-WD | 1872,31 | **64,24 ± 7,36** |
| GSO-$S_k$-WD | 2343,07 | 61,52 ± 8,07 |
| GSO-$H_k$-WD | 2389,07 | 62,44 ± 7,03 |

Os resultados mostram que o CGSO-$H_k$-WD obteve os melhore resultados entre os métodos avaliados para dois conjuntos de dados testados, o CGSO-$H_k$-WD para uma base e o CGSO-WD também para uma base. As abordagens baseadas no GSO obtiveram melhor desempenho do que GSO tradicional, apesar desta melhora não ter sido estatisticamente confirmada para a base vidro.

## 5 Conclusão

Neste artigo, apresentamos uma nova abordagem de aprendizagem baseada na hibridização do algoritmo LM com o método GSO baseado na cooperação, nomeadamente CGSO. LM usa a mecânica dos mínimos quadrados para determinar analiticamente os pesos de saída e o algoritmo GSO para otimizar a entrada de pesos e bias ocultas. O desempenho dos algoritmos foi avaliado com conjuntos de dados de classificação clássicos do UCI Machine Learning Repository. Os resultados experimentais mostram que as abordagens híbridas propostas obtiveram melhor generalização do que o GSO original e o algoritmo LM para todos os conjuntos de dados testados. Para a maioria dos casos, as técnicas baseadas no GSO superaram o GSO tradicional e LM, de acordo com o teste de avaliação ANOVA. Em trabalhos futuros podemos considerar outras bases de dados e diversificar os tipos de problemas, bem como inserir novos nodos na camada escondida durante o processo de treinamento.

## 6 Referência